%% file: ms.tex
\documentclass[utf8]{FrontiersinHarvard} 
\usepackage{url,hyperref,lineno,microtype,subcaption}
\usepackage{datetime, fmtcount, etoolbox, fcprefix}
\usepackage{threeparttable}
\usepackage[onehalfspacing]{setspace}
\usepackage{color,soul}
\usepackage{tabularx,colortbl,multirow,float}

\usepackage{enumitem} 

\usepackage{flushend}
\usepackage{mdwlist}
\usepackage{pifont}
\usepackage{float}
\usepackage{hyperref}       
\usepackage{url}            
\usepackage{booktabs}       
\usepackage{amsfonts}       
\usepackage{nicefrac}       
\usepackage{microtype}      
\usepackage{xcolor}         
\usepackage{comment}
\usepackage{multirow}
\usepackage[dvipdfmx]{graphicx}
\usepackage{tabularx}
\usepackage{setspace}
\usepackage{hyperref}
\hypersetup{
    colorlinks=true,
    urlcolor=blue,
}
\usepackage{amsthm}
\usepackage{csquotes}
\usepackage{soul}

\def\keyFont{\fontsize{8}{11}\helveticabold }
\def\firstAuthorLast{Mohit Prabhushankar {et~al.}} 
\def\Authors{Mohit Prabhushankar\,$^{*}$, Ghassan AlRegib\,}
\def\Address{ Omni Lab for Intelligent Visual Engineering and Science (OLIVES), Georgia Institute of Technology, Electrical and Computer Engineering, Atlanta, GA, USA}
\def\corrAuthor{Mohit Prabhushankar}

\def\corrEmail{mohit.p@gatech.edu}

\theoremstyle{definition}
\newtheorem{definition}{Definition}[section]

\begin{document}
\onecolumn 

\begin{basedescript}{\desclabelstyle{\pushlabel}\desclabelwidth{6em}}

\item[\textbf{Citation}]{M. Prabhushankar and G. AlRegib, "Stochastic Surprisal: An inferential measurement of Free Energy in Neural Networks," \emph{Frontiers in Neuroscience - Perception Science}, 17, 2023.}

\item[\textbf{Review}]{Data of Initial Submission : 22 April 2022 \\ Date of First Revision : 13 Dec 2022 \\ Date of Second Revision : 08 Feb 2023 \\ Date of Acceptance: 09 Feb 2023}

\item[\textbf{Codes}]{\url{https://github.com/olivesgatech/Stochastic-Surprisal}}

\item[\textbf{Copyright}]{\textcopyright 2023 Prabhushankar and AlRegib. This is an open-access article distributed under the terms of the Creative Commons Attribution License (CC BY). The use, distribution or reproduction in other forums is permitted, provided the original author(s) or licensor are credited and that the original publication in this journal is cited, in accordance with accepted academic practice. No use, distribution or reproduction is permitted which does not comply with these terms. }

\item[\textbf{Contact}]{\href{mailto:mohit.p@gatech.edu}{mohit.p@gatech.edu}  OR \href{mailto:alregib@gatech.edu}{alregib@gatech.edu}\\ \url{https://ghassanalregib.info/} \\ }

\end{basedescript}
\thispagestyle{empty}
\newpage
\clearpage
\setcounter{page}{1}

\onecolumn
\firstpage{1}

\title[Stochastic Surprisal in Neural Networks]{Stochastic Surprisal: An inferential measurement of Free Energy in Neural Networks}

\author[\firstAuthorLast ]{\Authors} 
\address{} 
\correspondance{} 

\extraAuth{}

\maketitle
\begin{abstract}

This paper conjectures and validates a framework that allows for action during inference in supervised neural networks. Supervised neural networks are constructed with the objective to maximize their performance metric in any given task. This is done by reducing free energy and its associated surprisal during training. However, the bottom-up inference nature of supervised networks is a passive process that renders them fallible to noise. In this paper, we provide a thorough background of supervised neural networks, both generative and discriminative, and discuss their functionality from the perspective of free energy principle. We then provide a framework for introducing action during inference. We introduce a new measurement called stochastic surprisal that is a function of the network, the input, and any possible action. This action can be any one of the outputs that the neural network has learnt, thereby lending \emph{stochasticity} to the measurement. Stochastic surprisal is validated on two applications: Image Quality Assessment and Recognition under noisy conditions. We show that, while noise characteristics are ignored to make robust recognition, they are analyzed to estimate image quality scores. We apply stochastic surprisal on two applications, three datasets, and as a plug-in on twelve networks. In all, it provides a statistically significant increase among all measures. We conclude by discussing the implications of the proposed stochastic surprisal in other areas of cognitive psychology including expectancy-mismatch and abductive reasoning. 

\tiny
\keyFont{ \section{Keywords:} Free Energy Principle, Neural Networks, Stochastic Surprisal, Image Quality Assessment, Robust Recognition, Human Visual Saliency, Abductive Reasoning, Active Inference} 
\end{abstract}

\input{Sections/1_Introduction}

\input{Sections/2_Methods}

\input{Sections/3_Results}

\input{Sections/4_Discussion}


\bibliographystyle{Frontiers-Harvard}
\bibliography{references}

\end{document}

%% file: Sections/1_Introduction.tex
\section{Introduction}
\label{sec:Introduction}
The human visual system is the resultant of an evolutionary process influenced and constrained by the natural visual stimuli present in the outside environment~\citep{geisler2008visual,sebastian2017constrained}. The free energy principle is an over-arching theory that provides a mathematical framework for this evolutionary process~\citep{friston2009free}. The principle provides a theory of cognition that can unify and discuss relationships among fundamental psychological concepts such as memory, attention, value, reinforcement, and salience~\citep{friston2009free}. It decomposes the visual system into perception and action modalities and argues that the visual system is an inference engine whose objective is to perceive the outside environment as best as it can. If this perception is insufficient for making an inference, an action is taken to achieve the objective by influencing the outside environment. While the action is dependent on the type of inference that is to be made, perception is dependent on the natural visual stimuli. Hence, a study of the human visual system warrants a study of the patterns that it is sensitive to. Broadly, these patterns are classified under natural scene statistics~\citep{geisler2008visual}. Color, luminance, spatio-temporal structures and spectral residues are some statistics that are useful in performing fundamental visual tasks including image quality assessment~\citep{zhang2012sr}, visual saliency detection~\citep{hou2007saliency}, and object detection and recognition~\citep{sebastian2017constrained}.

Image quality assessment is the objective assessment of subjective quality of images. Visual saliency detection finds those regions in an image that attract significant human attention. Object recognition attempts to recognize any given object in an image. Methods like~\citep{hou2007saliency,murray2013low} use spectral residue to detect salient regions.~\cite{hou2007saliency} extend their spectral residue-based saliency detection algorithm to show that object detection is possible. The spectral residual concept is used in SR-SIM~\citep{zhang2012sr} and BleSS~\citep{temel2016bless} to utilize the frequency characteristics to quantify residuals for IQA. All three disparate applications share commonalities in their spectral residual statistics that are used to show comparable performance within each application. Hence, natural scene statistics and their governing visual system principles are building blocks of computational machine vision systems that attempt to mimic human perception.

One such a principle is the consistency in spatial structures that allows for a sparse set of convolutional kernels to represent natural scenes. Large-scale neural networks are built on this principle. Neural networks are empowered to mimic human vision by performing the same tasks as the human visual system including image quality assessment~\citep{temel2016unique}, visual saliency detection~\citep{sun2020implicit}, and object recognition~\citep{krizhevsky2012imagenet} among others. Recently the generalization capabilities of neural networks has led to their widespread adoption in a number of computational fields. Neural networks have produced state-of-the-art results on multifarious data ranging from natural images~\citep{krizhevsky2012imagenet}, computed seismic~\citep{shafiq2018towards, shafiq2018leveraging}, and biomedical images~\citep{prabhushankar2022olives, prabhushankar2021extracting}. In object recognition on Imagenet dataset~\citep{deng2009imagenet},~\cite{he2016deep} surpassed the top-$5$ human accuracy of $94.9\%$. In the application of image quality assessment,~\cite{bosse2017deep} extracted patch-wise distortion characteristics from images using deep neural networks before fusing them to obtain an objective quality score. The authors in~\cite{liu2017reduced} device a sparse representation-based entropic measure of quality that is inspired by the free energy principle. This is extended in~\cite{liu2019unsupervised} where the authors use the free energy principle as a plug-in on top of existing blind image quality assessment techniques. In both these works, free energy principle is seen as a technique that measures the disparity between an outside environment and the the expectation of that environment through some biologically plausible mechanism. Other existing works, including~\citep{zhai2011psychovisual, gu2014using}, quantify this disparity to estimate quality.

Hence, from the perspective of free energy principle, neural networks act as biologically plausible mechanisms to perceive the outside environment. This is done by supervising the networks to learn particular tasks.~\cite{prabhushankar2021contrastive} describe supervised learning as associative learning where a set of learned features is associated with any given class. This class can be an objective score in image quality assessment or an object class from recognition. The learned features are associated with a specific dataset and application, and are not easily transferable~\citep{temel2018cure}. A number of recent works including~\citep{temel2017cure, goodfellow2014explaining, hendrycks2019benchmarking} describe the fallibility of neural networks to adversarial noise and slight perturbations in data arising from acquisition or environmental errors. The feature representation space can be altered significantly by noise that is sometimes non-noticeable in data. This is in contrast with the spectral residual feature which is used to infer both object~\citep{hou2007saliency} and image quality~\citep{zhang2012sr, temel2016bless}.

We posit that these shortcomings of supervised neural networks are a resultant of neural networks exclusively utilizing the perception modality of free energy principle. In other words, the passivity of neural networks during inference leads to their non-robust nature. This view is corroborated by~\cite{demekas2020investigation} who identify three challenges in supervised learning. Firstly, they claim that neural networks lack an explicit control mechanism of incorporating prior beliefs into predictions. Secondly, neural networks train via a scalar loss function that does not allow for incorporating uncertainty in action. Lastly, neural networks do not perform any action during inference that would elicit changes in the input from the outside environment. 

In this paper, we tackle the above challenges by introducing a framework for action during inference. This is opposed to the free energy principle based works in~\cite{liu2017reduced, liu2019unsupervised} where the methodology does not require actions at inference. Based on the free energy principle, we treat any trained neural network as an inference engine. We define a quantity called \emph{stochastic surprisal} that is a function of a neural network's inference and some action performed on this inference. Reducing surprisal is generally seen as a single action that reduces the distributional difference between two quantities. However, during inference, we have access to only a single data point. We overcome this challenge by considering that all possible actions that the network can undertake are equally likely. The term \emph{stochastic} is derived based on this assumption of action-randomness. Stochastic surprisal acts on top of \emph{any} existing neural networks to address the challenge of passive inference. Existing neural networks can either be generative or discriminative. We evaluate stochastic surprisal on two applications including image quality assessment and robust object recognition. In image quality assessment, we evaluate our technique to assess the quality of distorted images at different levels of distortions. Similarly, in robust object recognition, we recognize distorted images when the original neural network is only trained on pristine images. In other words, we propose a concept that is able to assess the noise characteristics in images to assign objective quality, as well as ignore the same noise characteristics to robustly classify images. The contributions of this paper include,

\begin{enumerate}

    \item We unify the concepts of image quality assessment and robust object recognition. We show that the features that are extracted from neural networks simultaneously characterize the scene and context within the image for recognition as well as the noise perturbing it's quality.
    \item We term our features as \emph{stochastic surprisal} and relate them to the free energy principle. We provide a mathematical framework to extract stochastic surprisal from both discriminative and generative neural networks as a function of some action.
    \item We discuss the implications of our proposed method from an abductive reasoning as well as expectancy-mismatch perspective. Both these concepts lead to separate applications including context and relevance based contrastive visual explanations and human visual saliency detection.

\end{enumerate}

We first describe the free energy principle in Section~\ref{subsec:FPE}. The free energy principle is then related to neural networks in Section~\ref{subsec:FPE_NN} before describing stochastic surprisal. The generation of stochastic surprisal in generative and discriminative networks is described in Sections~\ref{subsec:Generative} and~\ref{subsec:Discriminative} respectively. Finally, the applications of image quality assessment and robust recognition and our methodology is discussed in Section~\ref{subsec:Experiments}. The results are provided in Section~\ref{sec:Results}. We further discuss the implications of the proposed stochastic surprisal on other cognitive concepts and conclude in Section~\ref{sec:Discussion}.

%% file: Sections/2_Methods.tex
\section{Theoretical Overview and Methodology}
\label{sec:Background}
In this section, we provide a thorough background of the free energy principle and its application in neural networks, both generative and discriminative. We then define and detail the framework for the extraction of stochastic surprisal. This is followed by the application of stochastic surprisal in image quality assessment and robust recognition. 

\subsection{Background}
\subsubsection{Free Energy Principle}
\label{subsec:FPE}
The Free Energy Principle (FEP) proposes a theory to explain the self-organizing capability of any intelligent and adaptive system~\citep{friston2009free}. FEP assumes the demarcation of a \emph{system} that exists in an \emph{environment} through a functional \emph{Markov Blanket}. The Markov Blanket~\citep{hipolito2021markov} provides statistical independence to the system from its environment, thereby imbuing the system with a sense of \emph{self}. A consequence of this separation is that the system only experiences the environment through the Markov Blanket based on a limited set of sensory inputs. These sensory inputs are used to create a generative model of the outside environment within the system. The system then performs a limited set of actions affecting the outside environment while updating its internal model of the outside environment. The FEP provides a mathematically concrete set of principles to bound the long-term entropy of the internal generative model that is confined in the set of all possible sensory inputs and its possible performative actions.~\cite{friston2019free} argues that the assumption of the Markov Blanket and the ensuing FEP is an overarching theory that provides a tool to study and explain self-organization at any spatio-temporal scale from infinitesimal quantum mechanics to generational biological evolution.

In this paper, we are interested in the FEP's application to visual processes related to the human brain. The applicability of FEP across concepts such as memory, attention, value, and reinforcement~\citep{friston2009free} is possible because of the central assumption that the~\emph{limited} sensory inputs from the outside environment to the brain are also~\emph{likely} sensory inputs. In other words, the human brain only allows for a \emph{limited} set of \emph{likely} encounters~\citep{demekas2020investigation}. The term \emph{likely} is a function of the expectation set by the internal generative model within the brain. Hence, the brain is considered to encode a bayesian recognition density that predicts the sensory inputs based on some hypothesis regarding their cause. This leads to the proposition that the brain is an inverse generative model where it expects to sense only a limited set of likely inputs from the environment. Any mismatch to this expectation is handled in two stages. Firstly, the internal model is updated with the mismatched sensory input to improve the \emph{perception}. Secondly, an action is performed to change the environment. This way, the environment and the model are made to fit each other by reducing the mismatched input. A mismatched input is typically termed as a \emph{surprising} event~\citep{buckley2017free}. Self-organization in the brain creates an imperative to minimize the~\emph{surprisal} of any event and the FEP provides a mathematical theory of this minimization by providing a tractable upper bound to the surprisal. Mathematically, average surprisal is the entropy of the distribution of all events. More \emph{unlikely} an event, more~\emph{surprisal} it creates in the internal model. The free energy decomposed using surprisal~\citep{demekas2020investigation} is given by,
\begin{equation}\label{eq:Text_FEP}
    \text{Free Energy} = \text{Divergence} + \text{Surprisal}.
\end{equation}
Here, divergence is the difference between the variables representing the outside environment that generate the sensory inputs and the variables in the internal generative model that mimic the outside world. 

\subsubsection{Free Energy Principle in Neural Networks}
\label{subsec:FPE_NN}
The assumption of the existence of an internal tractable generative model that is an inference engine has been adopted in the construction of early neural networks.~\cite{hinton1993autoencoders} describe the Helmholtz free energy that is used to construct autoencoders as agents that minimize the reconstruction cost and the code cost. The code cost is a function of the entropy of the probability distribution given a vector. In FEP, this code cost is the surprisal. Variational Autoencoders~\citep{kingma2019introduction} minimize Variational Free Energy (VFE) and consequently surprisal. VFE is a generalization of the Helmholtz free energy where the divergence of the approximate and true probabilities are minimized~\citep{gottwald2020two}. While the generative models of autoencoders lend themselves directly to the FEP, the discriminative models also train themselves using some variation of a loss function that resembles free energy. In this paper, we use both generative and discriminative models and we introduce them in terms of the free energy principle.  

\begin{figure}[t]
 	\centering
 	\includegraphics[width=1\columnwidth]{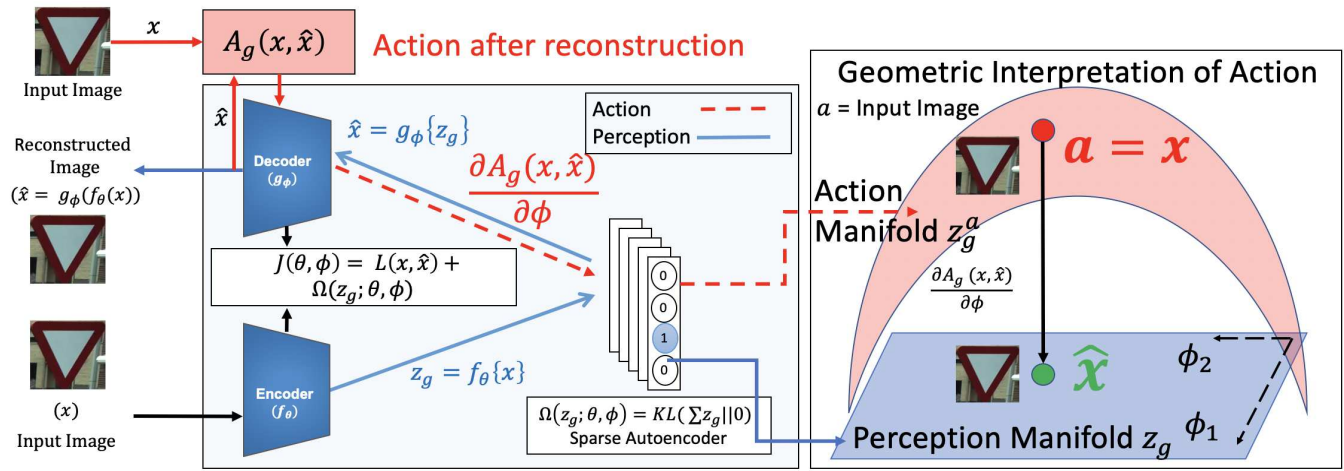}
 	\caption{Block diagram for perception (in blue) and proposed action (in red) for a sparse autoencoder. The image $x$ is taken from the CURE-TSR dataset~\citep{temel2017cure}. The training loss function is $J(\theta, \phi)$. The latent representation $z = f_\theta(\cdot)$ is $z_g$. The reconstructed image is shown as $\hat{x}$. This forms the perception pipeline. The action pipeline is shown in red where the action $\mathcal{A}_g$ is backpropagated through the decoder to the latent representation space. The learned blue perception representation space $z_g$ changes to the action space $z_g^a$ as a consequence of $\mathcal{A}_g$. This change is stochastic surprisal, given by $\frac{\partial \mathcal{A}_g(x, \hat{x})}{\partial \phi}$.}\label{fig:AE_BlockDiagram}
\end{figure}

\paragraph{Generative Networks}
\label{subsec:Generative}
In this section, we consider a general autoencoder as our generative model. An autoencoder is an unsupervised learning network which learns a regularized representation of inputs to reconstruct them as its output~\citep{hinton1993autoencoders, kwon2019distorted}. Since~\cite{hinton1993autoencoders}, a number of variations have been proposed to autoencoders to construct either application-specific or property-specific networks. These variations generally deal with constraining the latent representations learned by an autoencoder. For instance,~\cite{ng2011sparse} constrain the latent representation to be sparse, thereby constructing sparse autoencoders.~\cite{Kingma2013} constrain the latent representation to follow a Gaussian distribution. These are termed as variational autoencoders. These are two instances of property-specific autoencoders. Application-specific autoencoders include fully-connected networks used for image compression~\citep{gedeon1992progressive}, and convolutional autoencoders for image denoising~\citep{mao2016image}.

All these autoencoders consist of the same base architecture as shown in Fig.~\ref{fig:AE_BlockDiagram}. They consist of an encoder $f_\theta(\cdot)$, parameterized by $\theta$ to map inputs $x$ to a latent representation $z_g$. These latent representations $z_g$ are used to reconstruct the same input $\hat{x}$ using a decoder $g_\phi(\cdot)$. This operation is mathematically represented as,
\begin{equation}\label{eq:AEs}
    z = f_\theta (x) \quad \hat{x} = g_\phi (z) = g_\phi (f_\theta (x)),
\end{equation}
For a natural image input, $x \in \mathbb{R}^{H \times W \times C}$, where $H, W, C$ are height, width, channel of input image, respectively. The encoder and decoder are trained jointly by minimizing a loss function $J(\theta,\phi)$ defined as:
\begin{equation}\label{eq:AE_loss}
    J(\theta, \phi) = \mathcal{L}(x, g_{\phi}(f_{\theta}(x))) + \Omega(z_g; \theta, \phi),
\vspace{-1.5mm}
\end{equation}
where $\mathcal{L}$ is a reconstruction error which measures the dissimilarity between the input $x$, and the reconstructed image $\hat{x}$. $\Omega$ is a regularization term added to avoid overfitting the network to the training set and to imbue the required constraints. For a sparse autoencoder, $\Omega$ is an $l_1$ sparsity constraint. However, since the $l_1$ constraint is not differentiable, a practical solution for constructing this sparsity constraint is to use KL-Divergence on $z_g$. Specifically, the sum of $z_g$ is constrained to either zero or a very small value using a distance metric like KL-Divergence. This is shown in Fig.~\ref{fig:AE_BlockDiagram} in blue. 

During training, the network parameters, $\theta$ and $\phi$ are updated by backpropagating the gradients of $J(\theta, \phi)$ w.r.t. the parameters. The update rule is given by,
\begin{equation}\label{eq:Backprop}
     \theta' = \theta - \frac{\partial J(\theta, \phi)}{\partial \theta}, \quad  \phi' = \phi - \frac{\partial J(\theta, \phi)}{\partial \phi},
\end{equation}
The two gradients provide the change in the network parameters required to incorporate better perception capabilities as measured by the loss function $J(\theta,\phi)$.

Consider Eq.~\ref{eq:AE_loss} and compare this against the free energy decomposition in Eq.~\ref{eq:Text_FEP}. The $\mathcal{L}$ reconstruction error measures the divergence. The regularization is the surprisal. Technically, regularization prevents the network from reconstructing $x$ exactly. Hence, surprisal is \emph{added} in generative networks to make them generalizable. A thorough analysis of regularization for reconstruction and feature transfer of autoencoders to multiple tasks is provided in~\cite{prabhushankar2018semantically}. While regularization impacts the reconstruction negatively, it enhances the adaptability and usability of features for generalized tasks and test sets. 

\paragraph{Discriminative Networks}
\label{subsec:Discriminative}

\begin{figure}[t]
 	\centering
 	\includegraphics[width=1\columnwidth]{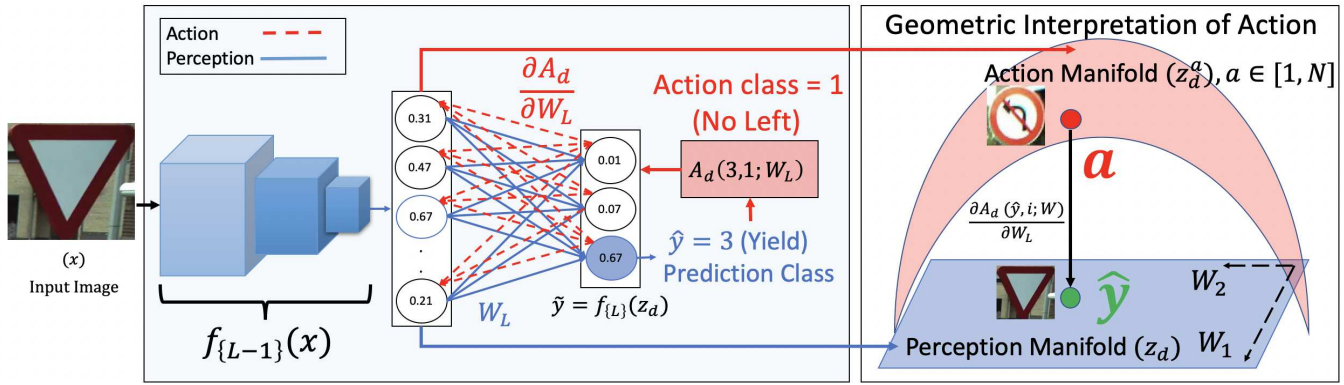}
 	\caption{Block diagram for perception (in blue) and proposed action (in red) for a classification network. The image $x$ is taken from the CURE-TSR dataset~\citep{temel2017cure}. The perception pipeline is shown in blue where the network assigns a class $3$ to $x$. The action pipeline is shown in red where the action $\mathcal{A}_d, a = 1$ is backpropagated through the final fully connected layer to the learned blue perception manifold $z_d$. $z_d$ changes to the action manifold $z_d^a$ as a consequence of $\mathcal{A}_d$. This change is stochastic surprisal, given by $\frac{\partial \mathcal{A}_d(\hat{y}, i;W)}{\partial W_L}$.}\label{fig:Discriminative_BlockDiagram}
\end{figure}

Discriminative networks are neural networks whose function is to assign labels to input data. While the required training data in generative networks are images $x \in \mathbb{R}^{H \times W \times C}$, the training data for discriminative networks are $(x, y)$, where $x \in \mathbb{R}^{H \times W \times C}$ and $y \in [1, N]$. Here, $y$ is an integer label assigned to $x$, ranging between $1$ and the total number of classes $N$. The goal of a discriminative network is to assign the label $y$, given $x$ at inference. The simplest discriminative network is an image classification network. Consider an $L$-layered network $f(\cdot)$ trained to classify images on a domain $\mathcal{X}$. For the task of classification, where $f(\cdot)$ is trained to classify between $N$ classes, the last layer is commonly a fully connected layer consisting of $N$ weights or filters. During inference, the representation space $z_d = f_{L-1}(x)$ after the first $(L-1)$ layers are projected independently onto each of the $N$ filters. The filter with the maximum projection is inferred as the class $\hat{y}$ to which $x$ belongs. Mathematically, $z_d$ and $\hat{y}$ are related as,
\begin{align}
    & z_d = f_{L-1}(x), \label{eq:feature}\\
    \tilde{y} = \operatorname*{arg\,max} & (W_L^T z_d + b_L), \quad \hat{y} = \operatorname*{arg\,max} (\tilde{y}) \label{eq:Filter}\\
    \forall  W_L\in \Re^{d_{L-1}\times N}, & z\in \Re^{d_{L-1}\times 1}, b_L\in \Re^{N\times 1}, \tilde{y} \in \Re^{N\times 1}, \hat{y} \in [1,N],
\end{align}
where $W_L$ and $b_L$ are the parameters of the final fully connected layer. Note our choice of the variable $z_d$. This is a similar variable that is used to denote the latent representation in Eq.~\ref{eq:AEs}. Similar to the decoder $g_\phi(\cdot)$ acting on $z_g$ in generative networks, we have the final fully connected layer $W_L$ and $b_L$ acting on $z_d$. This forms the perception pipeline that classifies $x$ as $\hat{y}$. This is shown in blue in Fig.~\ref{fig:Discriminative_BlockDiagram}.

Training an image classification technique requires a loss function $J(\hat{y},y; \theta)$, where $\theta$ are the network parameters and $(x,y)$ are the image-label pairs required for training. A common choice of $J(\cdot)$ is the cross-entropy loss. Considering $\sigma(\tilde{y})$ to be the softmax probability distribution of the output vector from $f(\cdot)$, the cross-entropy loss interms of KL-Divergence and entropy can be expressed as,
\begin{equation}\label{eq:Classifier}
    J(\cdot) = \text{KL}(y||\sigma(\tilde{y})) - \sum_{i=1}^{N} (\sigma(\tilde{y_i}))ln(\sigma(\tilde{y_i})).
\end{equation}
Here, $\text{KL}(||)$ refers to the KL-divergence between the probability output of the network and the label vector $y$ expressed as a one-hot probability distribution. Notice the similarity between Eqs.~\ref{eq:Text_FEP} and~\ref{eq:Classifier}. The divergence in the FEP is the KL divergence and the surprisal is the entropy given by the second term in Eq.~\ref{eq:Classifier}. Unlike the generative networks, surprisal is not introduced into the network. Rather, the existing surprisal is minimized. A number of foundational works in FEP~\citep{friston2009free, friston2019free} use the entropy of a distribution to describe free energy. The network is then trained by backpropagating the errors w.r.t $\theta$ similar to Eq.~\ref{eq:Backprop}.

\paragraph{Terminologies}
Before describing our contributions, we summarize a few key terminologies that are extensively used within the FEP setup and how they relate to neural networks.

\noindent\textbf{External state of the world} $\mathcal{X}$ is the observed distribution of the outside world and each $x \in \mathcal{X}$ is an instance of this distribution. When describing discriminative systems, data is denoted as $(x,y)$ where $x$ is the data point and $y$ is its label. When dealing with generative models, data is $x$ only. When there is some distortion associated with the outside environment, the sampled data is $x'$ and the distribution is $\mathcal{X}'$. We will see $\mathcal{X}'$ and $x'$ in IQA and recognition experiments when input data are distorted by noise.

\noindent\textbf{System} A neural network $f(\cdot)$ trained on a distribution $\mathcal{X}$. A trained system is one that does not take in any external inputs to change or update its weights. We consider that a trained system is at NESS density. For a discriminative network, $f(\cdot)$ is the entire system and its training data is denoted by $(x,y)$. For a generative network, $f_\theta(\cdot)$ is an encoder trained to produce a latent representation space $z_g$ given data denoted by $x$ and $g_\phi(\cdot)$ is the decoder trained to reconstruct the image given a latent representation $z_g$.

\noindent\textbf{Markov Blanket} The part of the system that produces the latent representation $z$. In a generative system the markov blanket is the encoder $f_\theta(\cdot)$ and in a discriminative system, the markov blanket is the initial part of the network from Eq.~\ref{eq:feature}, $f_{L-1}(\cdot)$.  

\noindent\textbf{Internal State of the system} Let $z$ denote the internal state of the latent representation within a system. Given a generative network, the latent representation after the encoder, $z_g = f_\theta(x)$ is the internal state. Given a discriminative network, the internal state is $z_d = f_{L-1}(x)$. The internal states of both the networks are interchangeably referred to as latent representations or as perception manifolds. Note that similar to external state, if an input $x$ is distorted to $x'$, its internal state is also distorted and we will use either $z'_d$ or $z'_g$ to denote the internal state of the system. Given any action, $a$, the internal state shifts to $z^a$ to accommodate this action without necessarily changing $x$. All these states are shown in Figs.~\ref{fig:AE_BlockDiagram} and~\ref{fig:Discriminative_BlockDiagram}.

\subsection{Stochastic Surprisal}
\label{subsec:Stochastic}
During inference, the networks are passive. As discussed in Section~\ref{sec:Introduction} and noted by~\cite{demekas2020investigation}, there is no mechanism to include a non-scalar surprisal that allows for an action during inference. In this paper, we alleviate this challenge by defining a new quantity called \emph{stochastic surprisal} as a function of a hypothetical action. Consider the differences in the existing definitions of surprisal. In generative networks from Eq.~\ref{eq:AE_loss}, surprisal is the induced regularization that prevents overfitting and creates specific constraints for a latent representation $z_g$. In discriminative networks from Eq.~\ref{eq:Classifier}, surprisal is the entropy of the network's predicted distribution obtained from a linear combination on $z_d$. While the surprisal in Eq.~\ref{eq:Text_FEP} deals with bounding the system's surprise of the distributional divergence between the internal model and external environment, the regularization-based and entropy-based definitions provide a mathematically-tractable definition in neural networks. In this paper, we provide a new mathematically-tractable definition of surprisal that is inherently a function of an action $\mathcal{A}$ and its effect on the network. A formal definition is provided first.

\begin{definition}[Stochastic Surprisal]\label{def:stochastic_surprisal}
Given a trained neural network $f_\theta(\cdot)$ parameterized by $\theta$, the gradient change $\frac{\partial \mathcal{A}}{\partial \theta}$ with respect to the network parameters for all possible actions $\mathcal{A}$ from the perspective of $f_\theta(\cdot)$ is termed stochastic surprisal.
\end{definition}

Stochastic surprisal measures the change required in the 
trained perception network to measure any given action $\mathcal{A}$. It is stochastic since it does not measure the divergence between distributions but rather a single data point's influence on the network. It is a non-scalar value that acts on the network parameters according to Eq.~\ref{eq:Backprop}. Note that we do not actually update the network. Rather, we only measure the network update and use it as a surprisal quantity. This update is possible based on some action all of which are considered equally likely. A thorough discussion of the naming is provided in Section~\ref{subsec:Terminology}.

\subsubsection{Action and Stochastic Surprisal}
\vspace{-1.5mm}
\label{subsec:Gradient Generation}
Action is a function of any application. We first define it in a general fashion for generative and discriminative networks. In Section~\ref{subsec:Experiments}, we define it specifically for image quality assessment and robust recognition. 
\paragraph{Generative Networks}
The action in generative networks is straightforward. Given an image $x$ and its reconstructed image $\hat{x}$, the possible action is to change the weight parameters in a way that reduces the disparity between $x$ and $\hat{x}$. In this paper, we quantify this disparity as the Mean Square Error given by $\lVert x -\tilde{x} \rVert_2^2$. However, as described in Section~\ref{subsec:Generative}, the surprisal is present in the regularization terms. Hence, any action performed has to account for this surprisal. In this paper, we use the elastic net regularization. The overall action that induces a change in the network is given by,
\begin{equation}\label{eq:Action_Gen}
    \mathcal{A}_g = \lVert x -\hat{x} \rVert_2^2 + \beta \sum_{j=1}^{h} {\rm KL}(z_j || \hat\rho_j) + \lambda\lVert W \rVert_2^2.
\end{equation}
where $\mathcal{A}_g$ is a generative action. $\lVert x -\hat{x} \rVert_2^2$ is the MSE loss function, and $\lVert W \rVert_2^2$ is the regularization on the weights. $\sum_{j=1}^{h} {\rm KL}(z_j || \hat\rho_j)$ is the sparsity constraint denoted as the divergence between the latent representation and some small value $\hat{\rho}_j, j \in [1,h]$ where $h$ is the size of the latent representation. By minimizing the KL divergence, the latent variables $z_j, j \in [1,h]$ are made sparse. $\beta$ and $\lambda$ are hyperparameters controlling the regularization.

Stochastic surprisal is the the gradient of this action $\mathcal{A}_g$ with respect to the decoder weights. The action pipeline along with the stochastic surprisal generation is shown in Fig.~\ref{fig:AE_BlockDiagram} in red. At inference, a test image is passed through a trained network and reconstructed. The action from Eq.~\ref{eq:Action_Gen} is calculated and backpropagated to the latent representation space $z_d$. The change, measured as the gradients, creates a change in $z_d$ and the new action manifold is termed $z_d^a$. A toy example of the geometric interpretation of this change is also shown Fig.~\ref{fig:AE_BlockDiagram}. The blue perception manifold $z_g$ that reconstructs $\hat{x}$ is acted on by $\mathcal{A}_g$ to obtain a new red action manifold $z_d^a$. The decoder can use this space to reconstruct $x$ exactly. In Section~\ref{sec:Results}, we show how these generated gradients can be used as features for image quality assessment. Note that we keep the perception pipeline as is and make no changes to the training process.

\paragraph{Discriminative Networks}
The action $\mathcal{A}_d$ in discriminative networks is more involved than generative networks. While in generative networks, the possible action is to reconstruct the image with higher fidelity, in discriminative networks, the action can take any one of $N$ outcomes. At inference, discriminative networks are given an image $x$ and asked to predict its label $y$. Assuming that $\hat{y}$ is the prediction, the action we use to elicit change in the network parameters is by backpropagating an action class $a$ in the loss function $J(\hat{y},a;W), a \in [1,N]$.
\begin{equation}\label{eq:Action_Dis}
    \mathcal{A}_d = \lVert a_i -\tilde{y} \rVert_2^2, i \in [1,N].
\end{equation}
Here $a_i$ is the action class defined as a Kronecker delta function given by,
\begin{equation}\label{eq:Kronecker}
    a_{i} =
    \begin{cases}
            1, &         \text{if } i=\text{class},\\
            0, &         \text{otherwise} 
    \end{cases}
\end{equation}
There is no regularization added to the discriminative action since the probability distribution $\sigma(\tilde{y})$ derived from $\tilde{y}$ is a function of its surprisal entropy. Note that we use an MSE function for $\mathcal{A}_d$ in Eq.~\ref{eq:Action_Dis} similar to $\mathcal{A}_g$ from Eq.~\ref{eq:Action_Gen}. An important difference between Eqs.~\ref{eq:Action_Gen} and~\ref{eq:Action_Dis} is the number of possible actions. In discriminative networks that classify between $N$ classes, there are $N$ possible $i$ in Eq.~\ref{eq:Action_Dis}. Hence, there are $N$ possible actions $\mathcal{A}_d$ and $N$ possible surprisals $\frac{\partial \mathcal{A}_d^i}{\partial W_L}, \forall i \in [1,N]$. The action pipeline for discriminative network for a toy example where the predicted class is $3$ and the action class is $1$ is shown in Fig.~\ref{fig:Discriminative_BlockDiagram} in red. The surprisals are the red gradients from the final fully connected layer. We also show the geometric interpretation of a given action on the learned representation space $z_d$. The blue perception manifold is acted upon by $\mathcal{A}_d^1$ through $\frac{\partial \mathcal{A}_d^1}{\partial W_L}$ to obtain the red action manifold. Note that there are $N$ such possible red $z_d^a$ due to the $N$ possible actions. This idea of $N$ separate gradients to characterize data is not new. In~\cite{settles2007multiple}, the authors construct positive and negative instance labels for a given input $x$ in a binary decision setting. This is done to quantify uncertainty in an active learning setting. In this paper, we extend this characterization to $N$-label settings and use the image-label pairs to extract stochastic surprisal from the network. 

Notice the difference in the definitions of action. In FEP, the generative model acts on the outside world creating a change that reduces its surprisal. Our definition in Eq.~\ref{eq:Action_Dis} is the same one that is used in I-FGSM~\citep{goodfellow2014explaining} adversarial generation technique. Eq.~\ref{eq:Action_Dis} is continuously applied and a gradient w.r.t. the input, i.e. $\frac{\partial \mathcal{A}_d}{\partial x}$, is added to $x$ until the prediction changes adversarially. Changing the input would be a true action from the FEP sense. However, in this paper, we do not explicitly change the outside world or $x$. Rather, we measure the effect of such a change on the network using $\frac{\partial \mathcal{A}_d}{\partial W_L}$ without making said change.

\subsection{Methodology}
\label{subsec:Experiments}
We validate the effectiveness of stochastic surprisal during inference on two applications: Image Quality Assessment (IQA) and Robust Classification. The action gradients, $\frac{\partial \mathcal{A}}{\partial \phi}$ are used in two ways. The first approach is to use the surprisal gradients as error directions. This is done by projecting images with and without distortions onto the gradient space and comparing them. In this case, the surprisal acts as a measurement between the images and acts as a Full-Reference IQA metric. The second approach is to directly use surprisal gradients as feature vectors. The directional change caused by the actions is dependent on the network, the input and the action class. By keeping the network same across action classes, surprisal becomes a characteristic of the data. This approach is explored for the application of robust classification. 

\begin{figure}[t]
 	\centering
 	\includegraphics[width=1\columnwidth]{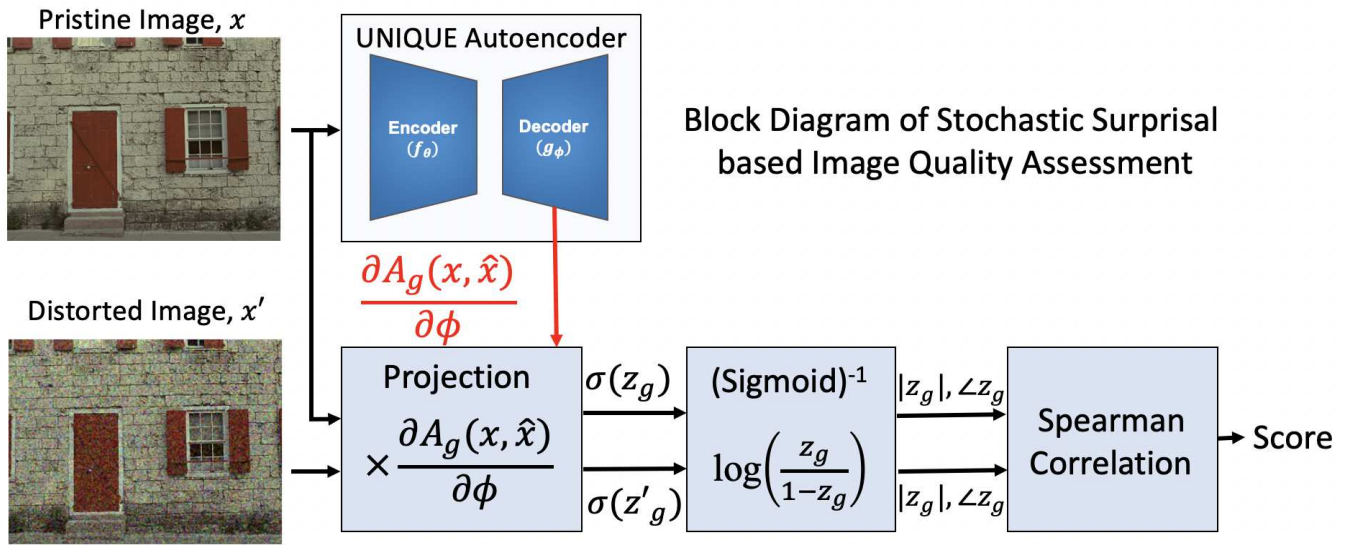}
 	\caption{Block diagram of the proposed framework of IQA as a plug-in on top of~\cite{temel2016unique}.}\label{fig:IQA_BlockDiagram}
\end{figure}

\subsubsection{Image Quality Assessment}
Image quality assessment is a field of image processing that objectively estimates the perceptual quality of a degraded image. Multiple methods have been proposed to predict the subjective quality of images~\citep{ponomarenko2011modified,wang2004image,wang2003multiscale,wang2011information,sampat2009complex,zhang2012sr,zhang2011fsim,mittal2012no,Temel_SPIC_2018, prabhushankar2017generating, prabhushankar2018semantically}. All these methods extract structure related hand-crafted features from both reference and distorted images and compare them to predict the quality. Recently, machine learning models directly extract features from images~\cite{temel2016unique, prabhushankar2017ms, bosse2017deep}. The authors in~\citep{bosse2017deep} propose to do so in either the presence or absence of the original pristine image. In~\cite{ma2021blind}, the authors propose a free energy inspired technique to predict the quality. They use a Generative-Adversarial Network as the base perception module and an additional CNN to model content and degradation dependent characteristics. In this paper, we approach the action module in FEP as a function of the perception module itself. We do so by extracting stochastic surprisal from the same perception network. Hence, our method acts as a plug-in on top of existing quality estimators. In this paper, we show quantitative results by plugging-in on top of UNIQUE~\citep{temel2016unique} and qualitative results on top of~\cite{bosse2017deep}. We first describe and motivate the usage of UNIQUE for quantitative results.

\noindent\textbf{UNIQUE:} We choose UNIQUE as the base technique since it follows the generative process described in Section~\ref{subsec:Generative} and Fig.~\ref{fig:AE_BlockDiagram}. This allows for the generation of stochastic surprisal from Eq.~\ref{eq:AE_loss} based on the Action in Eq.~\ref{eq:Action_Gen}. The authors in~\cite{temel2016unique} train a sparse autoencoder with a one layer encoder and decoder and a sigmoid non-linearity on $100,000$ patches of size $8\times 8 \times 3$ extracted from ImageNet~\citep{deng2009imagenet} testset. The autoencoder is trained with MSE reconstruction loss. This network is $f(\cdot)$ from Eq.~\ref{eq:AE_loss}. UNIQUE follows a full reference IQA workflow which assumes access to both reference and distorted images while estimating quality. The reference and distorted images are converted to YGCr color space and converted to $8 \times 8 \times 3$ patches. These patches are mean subtracted and ZCA whitened before being passed through the trained encoder. The activations of all reference patches in the latent space are extracted and concatenated. Activations lesser than a threshold of $0.025$ are suppressed to $0$. The choice of threshold $0.025$ is made based on the sparsity coefficient used during training. Similar procedure is followed for distorted image patches. The suppressed and concatenated features of both the reference and distorted images are compared using Spearman correlation. The resultant is the estimated quality of the distorted image. 

\paragraph{Proposed Methodology}
\label{subsec:IQA_Experiments}
We provide the block diagram for the proposed methodology in Fig.~\ref{fig:IQA_BlockDiagram}. Both the pristine and distorted images go through the same pre-processing steps detailed in UNIQUE~\citep{temel2016unique} and are projected onto the stochastic surprisal gradients of the decoder. The gradients $\frac{\partial \mathcal{A}_g}{\partial \phi}$ are extracted by backpropagating Eq.~\ref{eq:Action_Gen}. In this paper, we use the same hyperparameters $\beta = 3$, $\lambda = 3e^{-3}$, and $\rho_j = 0.035$ as used in~\cite{temel2016unique}. Once projected, the resultant is passed through an inverse sigmoidal layer to obtain the latent representation. Note that the latent representation is $z_g$ for the pristine image and $z'_g$ for the distorted image. Once passed through the inversion layer, both the magnitude and phase of each latent representation is concatenated and their spearman correlation coefficient is taken to estimate the quality score of the image. 

\subsubsection{Robust Classification}
The goal is to characterize an image $x$ using all $N$ actions. Consider an image $x$ whose class as predicted by $f_\theta(\cdot)$ is $\hat{y}$. Stochastic surprisal of $x$ against class $1$ is provided by backpropagating a loss between $\hat{y}$ and $1$ and obtaining corresponding gradients. The gradient is proportional to $\mathcal{A}_d(\hat{y},1;W_L)$, where $W$ is the weight parameters and $1$ is the action class. Specifically, it is $\nabla_{W_L} \mathcal{A}_d(\hat{y},1;W_L)$ for weights in layer $L$ and class $i \in [1,N]$. We backpropagate over all $N$ classes to obtain the overall surprisal features across all classes. The final feature, $r_x$ for an image $x$, is given by concatenating all individual features and $r_x$ is characteristic of image $x$. Hence,
\begin{equation}\label{eq:r_dis}
\begin{gathered}
    r_i = (\nabla_{W_L} \mathcal{A}_d(\hat{y},i;W_L))), \forall i \in [1,N], \\
    r_x = [r_1, r_2 \dots r_N].
\end{gathered}
\end{equation}

Given a trained feed-forward network $f(\cdot)$ and image $x$, we extract gradients using Eq.~\ref{eq:r_dis} which serve as our features. Gradients as features are used in diverse applications including visual explanations~\citep{selvaraju2017grad, prabhushankar2020contrastive, prabhushankar2021extracting}, adversarial attacks~\citep{goodfellow2014explaining}, anomaly detection~\citep{kwon2020backpropagated}, and image quality assessment~\citep{kwon2019distorted} among others. In this work, we use gradients as features to characterize data.

\input{Tables/Frameworks.tex}

\noindent\textbf{MLP ($\mathcal{H}(\cdot)$): }Once $r_x$ is obtained for all $N$ classes, the surprisal feature is now analogous to $z_d$ from Eq.~\ref{eq:feature}. However, $r_x$ is of dimensionality $\Re^{(N\times d_{L-1})\times 1}$ since it is a concatenation of $N$ gradients. To account for the larger dimension size, we classify $r_x$ by training an MLP $\mathcal{H}(\cdot)$ on top of $r_x$ derived from training data. In this paper, we use a simple three layered MLP as $\mathcal{H}(\cdot)$ with sigmoid activations. The exact structure of the MLP is dependent on $d_{L-1}$ of the base $f(\cdot)$ network and is given in Table~\ref{table:Architectures} for ResNets 18,34,50, and 101~\citep{he2016deep} that are considered in Section~\ref{sec:Results}.

\noindent\textbf{Training $\mathcal{H}(\cdot)$: }The concatenated $r_x$ features for all training data are extracted and normalized. $\mathcal{H}(\cdot)$ is trained on all training $r_x$ using the same training procedure as the perception network $f(\cdot)$. $\mathcal{H}(\cdot)$ is trained for $200$ epochs with SGD optimizer and cross-entropy loss, momentum $= 0.9$, weight decay $= 5e-4$, and adaptive learning rates of $0.1, 0.02, 0.004$ changed at epochs $60, 120, 160$ respectively.

\noindent\textbf{Testing using $f(\cdot)$ and $\mathcal{H}(\cdot)$:} During test time, the proposed framework operates in three steps. In step 1, as shown in Eq.~\ref{eq:Filter_H}, the given image passes through the perception network to provide a coarse estimation $\hat{y}$. In step 2, the stochastic surprisal features $r_x$ are extracted according to Eq.~\ref{eq:Filter_mid} and concatenated. In step 3, $r_x$ is normalized and passed through the MLP $\mathcal{H}(\cdot)$ to obtain the final prediction $\Tilde{y}$. This is shown in Eq.~\ref{eq:Filter_H_F}. 
\begin{align}
    &\hat{y} = \operatorname*{arg\,max} f(x),\label{eq:Filter_H}\\
    &r_x = [(\nabla_{W_L} \text{MSE}(\hat{y}, \delta^i_{i})), \forall i \in [1,N]],\label{eq:Filter_mid} \\
    &\Tilde{y} = \mathcal{H}(r_x),\label{eq:Filter_H_F}
\end{align}
Note that we substituted $\mathcal{A}_d$ in Eq.~\ref{eq:Filter_mid} with the MSE formulation of action from Eq.~\ref{eq:Action_Dis}.

%% file: Tables/Frameworks.tex
\begin{table*}[t]
\begin{center}
\begin{small}
\begin{sc}
\caption{Structure of $\mathcal{H}(\cdot)$ for different ResNet architectures as $f(\cdot)$.}
\label{table:Architectures}
\begin{tabular}{l c} 
 \toprule
 Network $f(\cdot)$ & Structure of $\mathcal{H}(\cdot)$ - All layers separated by sigmoid \\
 \midrule
 ResNet-18,34 &  $640\times300-300\times100-100\times10$ \\  
 ResNet-50, 101 & $2560\times300-300\times100-100\times10$ \\ 
 \midrule
\end{tabular}
\end{sc}
\end{small}
\end{center}
\end{table*}

%% file: Sections/3_Results.tex
\section{Results}
\label{sec:Results}

\input{Tables/tab_IQA_All_Extended.tex}
\subsection{Image Quality Assessment}
\label{sec:IQA_Results}
We report the results of the our proposed method in comparison with commonly cited methods in this section. We first discuss the the datasets used for comparison as well as the evaluation metrics. We finally show the results in Table~\ref{tab:iqa_results_all} and discuss these results.

\noindent\textbf{Datasets} We compare our proposed quality estimation technique on three datasets - \texttt{MULTI-LIVE}~\citep{jayaraman2012objective}, \texttt{TID2013}~\citep{ponomarenko2015image}, and \texttt{DR IQA}~\citep{athar2023degraded}. We choose \texttt{MULTI-LIVE} and \texttt{TID2013} datasets for two reasons. Firstly, our proposed technique is a plug-in approach on top of an existing technique~\citep{temel2016unique}. Hence, it is imperative to compare against and show results on datasets that were used in~\cite{temel2016unique}. Secondly, the two datasets provide access to seven categories of distortion among five levels. This is useful in comparison against the recognition experiments discussed in Section~\ref{sec:Results_Classification} which follows a similar setup. The complex distortions can either be a combination of multiple distortions such as distortions generated in the MULTI-LIVE dataset~\cite{jayaraman2012objective} or the human visual system (HVS) specific peculiar distortions such as the ones presented in the \texttt{TID2013}~\citep{ponomarenko2015image} dataset. A more challenging scenario is presented in \texttt{DR IQA} dataset, where the authors conjecture a degraded reference setting for image quality assessment. In this setting, pristine images are unavailable as a reference. Instead, singly distorted images are used as reference to construct IQA metrics for multiply distorted images. In Table~\ref{tab:iqa_results_all}, we provide results for \texttt{DR IQA} dataset as DRv1 and DRv2 based on the author's division of the dataset. Each of DRv1 and DRv2 have $31,790$ multiply distorted images and $1,122$ singly distorted images. Additionally, this dataset does not have \emph{true} subjective quality scores from humans but is derived from a synthetic quality benchmark. This synthetic score uses existing Full Reference metrics for quality generation including some of comparisons in Table~\ref{tab:iqa_results_all}.

\noindent\textbf{Evaluation metrics}
The performance is validated using outlier ratio (consistency), root mean square error (accuracy), Pearson correlation (linearity), Spearman correlation (rank), and Kendall correlation (rank). Arrows next to each metric in Table~\ref{tab:iqa_results_all} indicate the desirability of a higher number ($\uparrow$) or a lower number($\downarrow$). Statistical significance between correlation coefficients is measured with the formulations suggested in ITU-T Rec. P.1401~\cite{ITU2012} and provided below each correlation coefficient. A \texttt{0} value corresponds to statistically similar performance, \texttt{-1} means the method is statistically inferior to proposed method, and \texttt{1} indicates that the method is statistically superior to proposed method. Two best performing methods for each metric are highlighted.

\noindent\textbf{Results} We compare our proposed stochastic surprisal-based UNIQUE against other image quality estimators based only on handcrafted features and perception pipeline in Table~\ref{tab:iqa_results_all}. These compared full reference estimators include PSNR-HA~\citep{ponomarenko2011modified}, SSIM~\citep{wang2004image}, MS-SSIM~\citep{wang2003multiscale}, CW-SSIM~\citep{sampat2009complex}, IW-SIM~\citep{wang2011information}, SR-SIM~\citep{zhang2012sr}, FSIM~\citep{zhang2011fsim}, FSIMc~\citep{zhang2011fsim}, PerSIM~\citep{temel2015persim}, CSV~\citep{Temel201692}, UNIQUE~\citep{temel2016unique}. We also compare against no reference metrics including BRISQUE~\citep{mittal2012no}, BIQI~\citep{moorthy2010two}, and BLIINDS2~\citep{saad2012blind}. All these techniques were also comapred against the base UNIQUE algorithm in~\cite{temel2016unique}. In addition to these, we compare against new estimators including COHERENSI~\citep{Temel_SPIC_2018} and SUMMER~\citep{Temel_SPIC_2018}. SUMMER beats UNIQUE among six of the ten categories. Note that we do not show results for BRISQUE, BIQI, and BLIINDS2 for \texttt{DR IQA} dataset since NR methods, that are generally trained on singly distorted images, exhibit a large performance gap on multiply distorted images~\citep{athar2023degraded}.

The proposed stochastic surprisal-based method plugs on top of UNIQUE and its results are provided under the last column in Table~\ref{tab:iqa_results_all}. It is always in the top two methods for \texttt{MULTI-LIVE} and \texttt{TID2013} datasets in all evaluation metrics. In particular, the proposed method achieves the best performance for all the categories except in OR and KRCC in \texttt{TID2013} dataset. UNIQUE, by itself, does not achieve the best performance for any of the metrics in \texttt{MULTI} dataset. However, the same network using the proposed gradient features significantly improves the performance and achieves the best performance on all metrics. For instance, UNIQUE is the third best performing method in \texttt{MULTI} dataset in terms of RMSE, PLCC, SRCC, and KRCC. However, the action-based features improve the performance for those metrics by $1.315$, $0.036$, $0.020$, and $0.023$, respectively and achieve the best performance for all metrics. This further reinforces the plug-in capability of the proposed method during inference. On \texttt{DR IQA} dataset, FSIM and FSIMc perform the best across all categories. The authors in~\cite{athar2023degraded} used FSIMc to construct DR IQA models. However, the proposed algorithm remains competitive among all evaluation metrics. The results are statistically significant in $53$ of the $78$ compared metrics across both DRv1 and DRv2. Note that a number of these compared FR-IQA metrics have been utilized to construct the synthetic ground truth quality scores. 

\subsection{Robust Classification}
\label{sec:Results_Classification}

Neural networks are sensitive to distortions in test that the network was not privy to during training~\citep{temel2017cure, temel2018cure, hendrycks2019benchmarking}. These distortions include image acquisition errors, environmental conditions during acquisition, transmission and storage errors among others. \texttt{CIFAR-10C}~\citep{hendrycks2019benchmarking} dataset consists of $19$ real world distortions each of which has five levels of degradation that distort the $10000$ images in \texttt{CIFAR-10} testset. Neural networks that use perception-only mechanics suffer performance accuracy drops on \texttt{CIFAR-10C}. Current techniques that alleviate the drop in perception-only accuracy require additional training data. The authors in~\cite{vasiljevic2016examining} show that finetuning or retraining networks using distorted images increases the performance of classification under the same distortion. However, performance between different distortions is not generalized well. For instance, training on gaussian blurred images does not guarantee a performance increase in motion blur images~\citep{geirhos2018generalisation}. Other proposed methods include training on style-transferred images~\citep{geirhos2018imagenet}, training on adversarial images~\citep{hendrycks2019benchmarking}, training on simulated noisy virtual images~\citep{temel2017cure}, and self-supervised methods like SimCLR~\cite{chen2020simple} that train by augmenting distortions. Augmix~\citep{hendrycks2019augmix} creates multiple chains of augmentations to train the base network. All these works require additional training data. Our proposed stochastic surprisal-based technique is a plug-in on top of any existing method that increases the base network's robustness to distortions without any need for new data.

\noindent\textbf{Experimental setup and dataset: }We use \texttt{CIFAR-10C}~\citep{hendrycks2019benchmarking} as our dataset of choice with all its $95$ distortions and degradation levels. ResNet-18,34,50, and 101~\citep{he2016deep} architectures are used as the base $f(\cdot)$ perception-only networks. These are trained from scratch on \texttt{CIFAR-10} dataset. Following the terminologies established in Section~\ref{sec:Background}, $\mathcal{X}$ is the training set of \texttt{CIFAR-10} and $\mathcal{X}'$ are the $19$ distorted domains in which the testing set of \texttt{CIFAR-10C} reside. Each of the $19$ corruptions have $5$ levels of distortions. Higher the level, higher is the distortion. The distortions include blur characteristics like gaussian blur, zoom blur, glass blur, and environmental distortions like rain, snow, fog, haze among others.

\input{Tables/Compare}
\noindent \textbf{Comparison against existing State of the Art Methods:}~In Table~\ref{tab:Results_Compare}, we compare the Top-$1$ accuracy between perception-only inference and our proposed stochastic surprisal-based inference. All the state-of-the-art techniques require additional training data - noisy images~\citep{vasiljevic2016examining}, adversarial images~\citep{hendrycks2019benchmarking}, self-supervision SimCLR augmentations~\citep{chen2020simple}, and augmentation chains~\citep{hendrycks2019augmix}. We term these perception-only techniques as $f'(\cdot)$ and we actively infer on top of them. For all $f'(\cdot)$ other than Augmix, the base network is a ResNet-18. For Augmix, we use WideResNet architecture following the authors in~\cite{hendrycks2019augmix}. Another commonly used robustness technique is to pre-process the noisy images to denoise them. Denoising $19$ distortions is, however, not a viable strategy assuming that the characteristics of the distortions are unknown. We use Non-Local Means~\citep{buades2011non} denoising and the results obtained are lower than the perception-only accuracy by almost $3\%$. However, the proposed technique on this model increases the results by $3.84\%$. We create untargeted adversarial images using FGSM attack~\citep{goodfellow2014explaining} and use them to train a ResNet-18 architecture. In the experimental setup of augmenting noise~\citep{vasiljevic2016examining}, we augment the training data of CIFAR-10 with six distortions provided by~\cite{temel2018cure} to randomly distort 500 CIFAR-10 training images to train $f'(\cdot)$. For all techniques, the proposed technique plugs on top of $f'(\cdot)$ and increases the accuracy to create robust networks. Note that in all the perception-only methods in Table~\ref{tab:Results_Compare}, we do not use the augmented data to train $\mathcal{H}(\cdot)$. The gain obtained is by creating actions on only the undistorted data. Even when the augmented network $f'(\cdot)$ gains on non-augmented $f(\cdot)$, the proposed technique plugs on top of $f'(\cdot)$ to provide additional gains.

\begin{figure}[!t]
\begin{center}
\minipage{\textwidth}%
  \includegraphics[width=\linewidth]{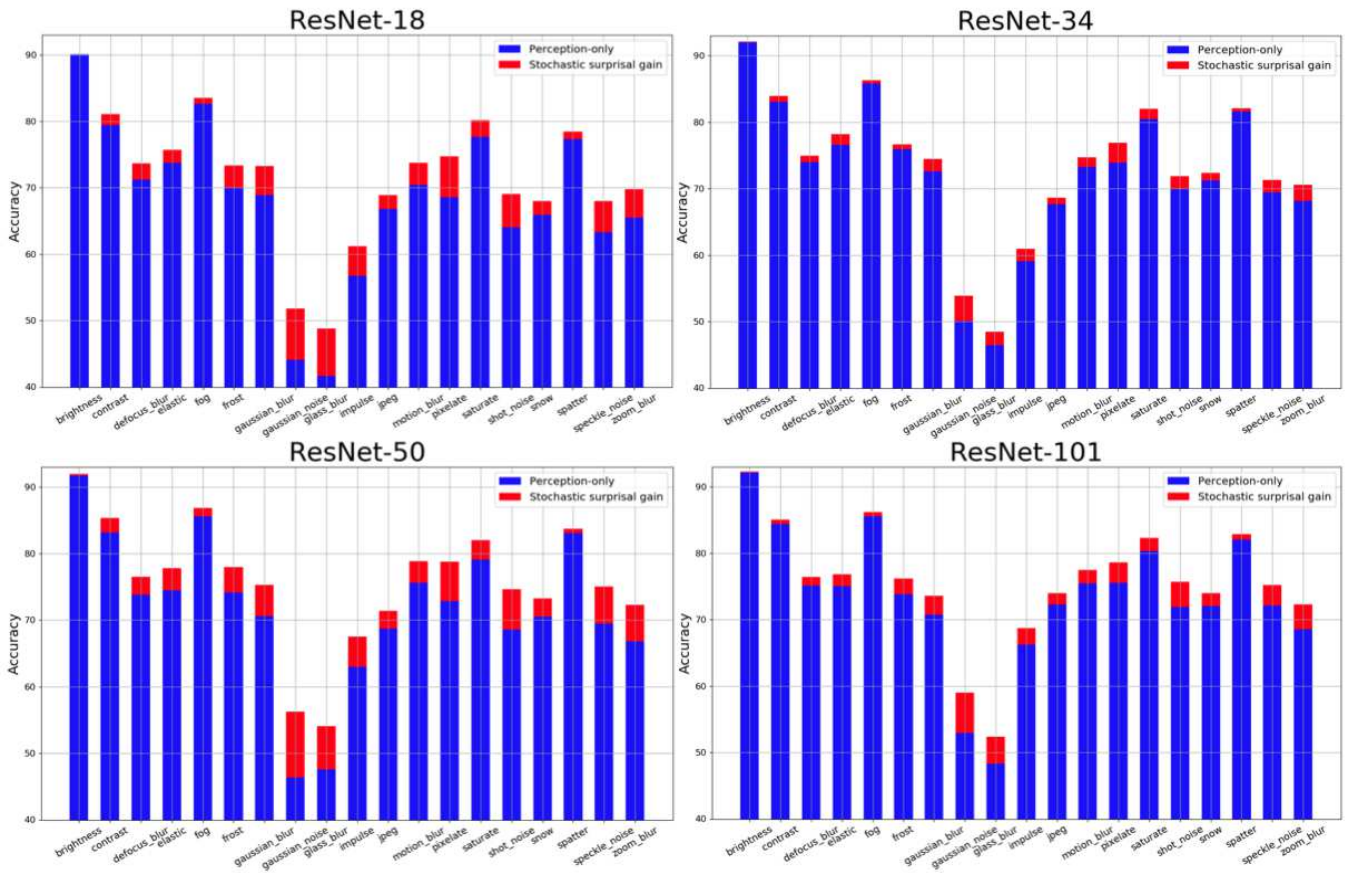}
\endminipage
\caption{Visualization of accuracy gains (in red) of using the proposed stochastic surprisal-based inference over perception-only inference (in red) on \texttt{CIFAR-10C} dataset~\citep{hendrycks2019benchmarking} for four networks across $19$ distortions.}\label{fig:Robustness_Distortion}
\end{center}
\end{figure}

\noindent\textbf{Analyzing distortion-wise accuracy gains: }The results of all four ResNet architectures for each of the $19$ distortions is shown in Fig~\ref{fig:Robustness_Distortion}. \texttt{X-Axis} in each plot shows $19$ distortions averaged over all $5$ distortion levels. \texttt{Y-Axis} shows Top-$1$ accuracy. The bars in blue show perception-only inference results and the red region in each bar represents the performance gain obtained by stochastic surprisal-based inference. There is an increase in performance across distortions and networks. In $9$ of the $19$ distortions, the proposed method averages $4\%$ more than its perception-only counterpart. These include gaussian blur, gaussian noise, glass blur, impulse noise, motion blur, pixelate, shot noise, speckle noise, and zoom blur. The highest increase is $8.22\%$ for glass blur. In $2$ of the distortions, brightness and saturate, the results increase by less than $0.4\%$ averaged over all levels. This is because of the statistics that the distortions affect. Distortions can change either the local or global statistics within images. Distortions like saturate, brightness, contrast, fog, and frost change the low level or global statistics in the image domain. Neural networks are actively trained to ignore such changes so that their effects are not propagated beyond the first few layers. Hence, gradients derived from the final fully connected layer do not capture the necessary changes required within $f(\cdot)$ to compensate for these distortions. Therefore, both the proposed and perception-only inference follow each other closely in distortions like brightness and saturate.

\begin{figure}[!t]
\begin{center}
\minipage{\textwidth}%
  \includegraphics[width=\linewidth]{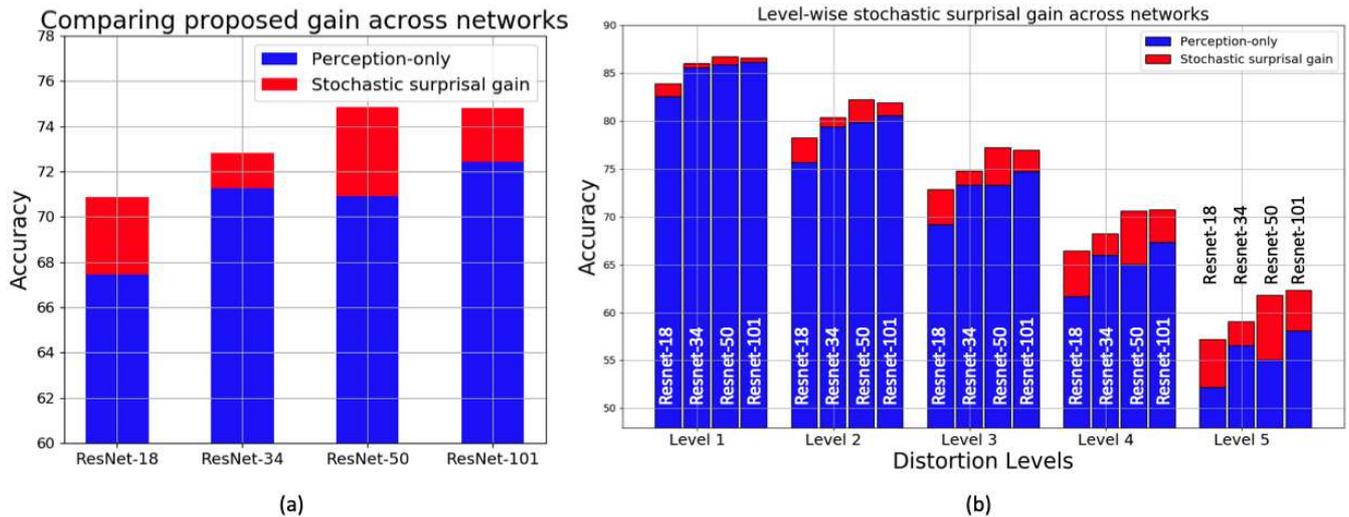}
\endminipage
\caption{Visualization of accuracy gains (in red) of using the proposed stochastic surprisal-based inference over perception-only inference (in blue) on \texttt{CIFAR-10C} dataset~\citep{hendrycks2019benchmarking} for four networks (a) averaged across $19$ distortions and $5$ levels (b) shown across $5$ levels of distortion.}\label{fig:Level}
\end{center}
\end{figure}

\noindent\textbf{Level-wise Recognition on CIFAR-10C: }
In Fig.~\ref{fig:Level}b, the proposed performance gains for the four networks are categorized based on the distortion levels. All $19$ categories of distortion on \texttt{CIFAR-10C} are averaged for each level and their respective perception-only accuracy and stochastic surprisal-based gains are shown. Note that the levels are progressively more distorted. Hence, level 1 distribution $\mathcal{X}'$ is similar to the training distribution $\mathcal{X}$ when compared to level 5 distributions. As the distortion level increases, the proposed method's accuracy gains also increase. This is because, with a larger distributional shift, more characteristic is the action required w.r.t. the network parameters. In Fig.~\ref{fig:Level}a, we show the distortion-wise and level-wise accuracy gains for each network. Note that, a stochastic surprisal-based ResNet-18 performs similarly to a perception-only ResNet-50.

%% file: Tables/tab_IQA_All_Extended.tex
\begin{table*}[htb!]
\tiny

\centering
\caption{Overall performance of image quality estimators.}
\label{tab:iqa_results_all}

\begin{threeparttable}

\rotatebox{90}{\begin{tabular}{p{1.5cm}p{0.75cm}p{0.75cm}p{0.75cm}p{0.75cm}p{0.75cm}p{0.75cm}p{0.75cm}p{0.75cm}p{0.75cm}p{0.75cm}p{0.75cm}p{0.75cm}p{0.75cm}p{0.75cm}p{1cm}p{1cm}p{1.1cm}}
\hline


\multirow{3}{*}{{\bf Databases }}                 & \bf PSNR  &\bf SSIM  &\bf MS  &\bf CW &\bf IW  &\bf SR  &\bf FSIM  &\bf FSIMc  &\bf BRIS  & \bf BIQI &\bf BLII &\bf Per &\bf CSV &\bf UNI &\bf COHER &\bf SUM &\bf Proposed \\
&\bf HA & &\bf SSIM  &\bf SSIM  &\bf SSIM  &\bf SIM  & &&\bf QUE & &\bf NDS2 &\bf SIM & & \bf QUE & \bf ENSI & MER & \\ 
\\ \hline



         
                & \multicolumn{17}{c}{\textbf{Outlier Ratio (OR, $\downarrow$)}}                                                                                                                                                                        \\ \hline
\textbf{MULTI}  
& 0.013 & 0.016 & 0.013 & 0.093 & 0.013 &\cellcolor{blue!10} \bf 0.000 & 0.018 & 0.016 & 0.067 & 0.024 & 0.078 & 0.004 &\cellcolor{blue!10} \bf 0.000 &\cellcolor{blue!10} \bf 0.000 & 0.031 &\cellcolor{blue!10} \bf 0.000 & \cellcolor{blue!10} \bf 0.000
\\                            
\textbf{TID13}  
&\cellcolor{blue!10} \bf 0.615 & 0.734 & 0.743 & 0.856 & 0.701 & 0.632 & 0.742 & 0.728 & 0.851 & 0.856 & 0.852 & 0.655 & 0.687 & 0.640 & 0.833 & \cellcolor{blue!10} \bf 0.620 & \cellcolor{blue!10} \bf 0.620
 \\ 
 \hline

                & \multicolumn{17}{c}{\textbf{Root Mean Square Error (RMSE, $\downarrow$)}}                                                                                                                                                                        \\ \hline

\textbf{MULTI}  
& 11.320 & 11.024 & 11.275 & 18.862 & 10.049 &\cellcolor{blue!10} \bf 8.686 & 10.866 & 10.794 & 15.058 & 12.744 & 17.419 & 9.898 & 9.895 & 9.258 & 14.806 & 8.212 & \cellcolor{blue!10} \bf 7.943\\

\textbf{TID13}
& 0.652 & 0.762 & 0.702 & 1.207 & 0.688 &\cellcolor{blue!10} \bf 0.619 & 0.710 & 0.687 & 1.100 & 1.108 & 1.092 & 0.643 & 0.647 & 0.615 & 1.049  & 0.630 & \cellcolor{blue!10} \bf 0.596
 \\ 

 \textbf{DRv1}
 & 16.19 & 17.11 & 16.17 & 17.18 & 14.02 &13.64 & \cellcolor{blue!10} \bf 12.98 & \cellcolor{blue!10} \bf 13.24 & - & - & - & 16.01 & 15.07 & 13.59 & 21.82  & 16.98 & 13.85
 \\ 

 \textbf{DRv2}
 & 16.47 & 16.42 & 15.76 & 17.48 & 14.04 &13.17 & \cellcolor{blue!10} \bf 12.82 & \cellcolor{blue!10} \bf 12.92 & - & - & - & 16.23 & 15.35 & 13.19 & 21.57  & 17.59 & 13.24 

\\
\hline

               & \multicolumn{17}{c}{\textbf{Pearson Linear Correlation Coefficient (PLCC, $\uparrow$)}}                                                                                                                                                                        \\ \hline

\multirow{2}{*}{{\bf MULTI}}                                  
& 0.801 & 0.813 & 0.803 & 0.380 & 0.847 & 0.888 & 0.818 & 0.821 & 0.605 & 0.739 & 0.389 & 0.852 & 0.852 &  0.872 & 0.622 &\cellcolor{blue!10} \bf  0.901 & \cellcolor{blue!10} \bf 0.908 \\                                          
& -1 & -1 & -1 & -1 & -1 & 0 & -1 & -1 & -1 & -1 & -1 & -1 & -1 & -1 & -1 & 0 & 

 \\

\multirow{2}{*}{{\bf TID13}}                               
& 0.851 & 0.789 & 0.830 & 0.227 & 0.832 & 0.866 & 0.820 & 0.832 & 0.461 & 0.449 & 0.473 & 0.855 & 0.853 & \cellcolor{blue!10} \bf 0.869 & 0.533  &  0.861 & \cellcolor{blue!10} \bf 0.877\\
& -1 & -1 & -1 & -1 & -1 & 0 & -1 & -1 & -1 & -1 & -1 & -1 & -1 & 0 & -1 & -1 & 
   \\ 

\multirow{2}{*}{{\bf DRv1}}                               
& 0.731 & 0.693 & 0.732 & 0.586 & 0.800 & 0.819 & \cellcolor{blue!10} \bf 0.833 & \cellcolor{blue!10} \bf 0.830 & - & - & - & 0.738 & 0.738 & 0.820 & 0.432  &  0.698 & 0.800\\
& -1 & -1 & -1 & -1 & 0 & 1 & 1 & 1 & - & - & - & -1 & -1 & 1 & -1 & -1 & 
   \\ 

\multirow{2}{*}{{\bf DRv2}}                               
& 0.709 & 0.702 & 0.738 & 0.521 & 0.799 & 0.826 & \cellcolor{blue!10} \bf 0.836 & \cellcolor{blue!10} \bf 0.833 & - & - & - & 0.720 & 0.720 & 0.825 & 0.417  &  0.658 & 0.815\\
& -1 & -1 & -1 & -1 & -1 & 0 & 1 & 1 & - & - & - & -1 & -1 & 0 & -1 & -1 & 
   \\ 
   
   \hline

\textbf{}      & \multicolumn{17}{c}{\textbf{Spearman's Rank Correlation Coefficient (SRCC, $\uparrow$)}}                                                                                                                                                                        \\ \hline

\multirow{2}{*}{{\bf MULTI}}                                          
& 0.715 & 0.860 & 0.836 & 0.631 & \cellcolor{blue!10} \bf 0.884 & 0.867 & 0.864 & 0.867 & 0.598 & 0.611 & 0.386 & 0.818 & 0.849 &  0.867 & 0.554  & \cellcolor{blue!10} \bf 0.884 & \cellcolor{blue!10} \bf 0.887 \\   
& -1 & 0 & -1 & -1 & 0 & 0 & 0 & 0 & -1 & -1 & -1 & -1 & -1 & 0 & -1 & 0 & 
\\

\multirow{2}{*}{{\bf TID13}}
& 0.847 & 0.742 & 0.786 & 0.563 & 0.778 & 0.807 & 0.802 & 0.851 & 0.414 & 0.393 & 0.396 &0.854 & 0.846 & \cellcolor{blue!10} \bf 0.860 & 0.649  & 0.856 & \cellcolor{blue!10} \bf 0.865 \\

& -1 & -1 & -1 & -1 & -1 & -1 & -1 & -1 & -1 & -1 & -1 & 0 & -1 & 0 & -1 & 0 &

  \\
\multirow{2}{*}{{\bf DRv1}}
& 0.739 & 0.702 & 0.738 & 0.760 & 0.798 & 0.807 & \cellcolor{blue!10} \bf 0.823 & \cellcolor{blue!10} \bf 0.820 & - & - & - &0.742 & 0.769 & 0.810 & 0.518  & 0.706 & 0.807 \\

& -1 & -1 & -1 & -1 & -1 & 0 & 1 & 1 & - & - & - & -1 & -1 & 0 & -1 & -1 &

  \\

\multirow{2}{*}{{\bf DRv2}}
& 0.720 & 0.705 & 0.738 & 0.755 & 0.795 & 0.809 & \cellcolor{blue!10} \bf 0.819 & \cellcolor{blue!10} \bf 0.816 & - & - & - &0.727 & 0.755 & 0.813 & 0.525  & 0.672 & \cellcolor{blue!10} \bf 0.816 \\

& -1 & -1 & -1 & -1 & -1 & -1 & 0 & 0 & - & - & - & -1 & -1 & 0 & -1 & -1 &

  \\

\hline

\textbf{}      & \multicolumn{17}{c}{\textbf{Kendall's Rank Correlation Coefficient (KRCC, $\uparrow$)}}                                                                                                                                                                        \\ \hline

\multirow{2}{*}{{\bf MULTI}}                                              
& 0.532 & 0.669 & 0.644 & 0.457 & \cellcolor{blue!10} \bf 0.702 & 0.678 & 0.673 & 0.677 & 0.420 & 0.440 & 0.268 & 0.624 & 0.655 &  0.679 & 0.399 & 0.698 & \cellcolor{blue!10} \bf 0.702\\                                         

& -1 & 0 & 0 & -1 & 0 & 0 & 0 & 0 & -1 & -1 & -1 & -1 & 0 & 0 & -1 & 0 & 

 \\  

\multirow{2}{*}{{\bf TID13}}                                          
& 0.666 & 0.559 & 0.605 & 0.404 & 0.598 & 0.641 & 0.629 & 0.667 & 0.286 & 0.270 & 0.277 &\cellcolor{blue!10} \bf 0.678 & 0.654 & 0.667 & 0.474   & 0.667 & \cellcolor{blue!10} \bf 0.677 \\
& 0 & -1 & -1 & -1 & -1 & -1 & -1 & 0 & -1 & -1 & -1 & 0 & 0 & 0 & -1 & 0 & 
\\ 

\multirow{2}{*}{{\bf DRv1}}                                          
& 0.534 & 0.505 & 0.537 & 0.559 & 0.597 & 0.609 & \cellcolor{blue!10} \bf 0.629 & \cellcolor{blue!10} \bf 0.626 & - & - & - & 0.537 & 0.563 & 0.609 & 0.357   & 0.503 & 0.605 \\
& -1 & -1 & -1 & -1 & 0 & 0 & 1 & 1 & - & - & - & -1 & -1 & 0 & -1 & -1 & 
\\ 

\multirow{2}{*}{{\bf DRv2}}                                          
& 0.517 & 0.509 & 0.539 & 0.553 & 0.595 & 0.613 & \cellcolor{blue!10} \bf 0.626 & \cellcolor{blue!10} \bf 0.623 & - & - & - & 0.525 & 0.594 & 0.613 & 0.342   & 0.475 & 0.616 \\
& -1 & -1 & -1 & -1 & -1 & 0 & 1 & 0 & - & - & - & -1 & -1 & 0 & -1 & -1 & 
\\ 

\hline
\end{tabular}
}
\end{threeparttable}
\vspace{-0.5cm}
\end{table*}

%% file: Tables/Compare.tex
\begin{table}[t!]
\centering
\scriptsize
\caption{Stochastic surprisal-based plug-in on top of existing robustness techniques.}
\vspace{2mm} 
\begin{sc}

\begin{tabular}{l c r} 
 \toprule
 Methods & & Accuracy  \\ 
 \midrule
 ResNet-18 & Perception-only & $67.89\%$ \\ 
 & Proposed & $\textbf{71.4}\%$ \\
 \midrule
 Denoising & Perception-only &$65.02\%$  \\
 & Proposed & $\textbf{68.86}\%$ \\
 \midrule
 Adversarial Train~\citep{hendrycks2019benchmarking} & Perception-only & $68.02\%$ \\
 & Proposed & $\textbf{70.86}\%$ \\
 \midrule
 SimCLR~\citep{chen2020simple} & Perception-only & $70.28\%$  \\ 
 & Proposed & $\textbf{73.32}\%$ \\
 \midrule
 Augment Noise~\citep{vasiljevic2016examining} & Perception-only & $76.86\%$  \\
 & Proposed & $\textbf{77.98}\%$ \\
 \midrule
 Augmix~\citep{hendrycks2019augmix} & Perception-only & $89.85\%$ \\
 & Proposed & $\textbf{89.89}\%$ \\
\midrule
\end{tabular}
\end{sc}
\label{tab:Results_Compare}\vspace{-3mm}
\end{table}

%% file: Sections/4_Discussion.tex
\section{Discussion}
\label{sec:Discussion}
We conclude this paper by considering the terminology of stochastic surprisal as well as some of the broader implications of the proposed technique. These include the abductive reasoning module and expectancy-mismatch hypothesis in cognitive science.

\subsection{Choice of the terminology of Stochastic Surprisal}\label{subsec:Terminology}
We motivate the terminology of \emph{stochastic surprisal} in two ways:
\begin{enumerate}
    \item As an analogy to~\emph{gradient descent} and \emph{stochastic gradient descent}: Gradient descent requires the gradients from the all available training data to update the weights. However, since this is computationally infeasible for large neural networks, stochastic gradient descent allows using a single training datapoint to estimate gradients, repeated across all data. In \emph{stochastic surprisal}, we use the single data point, available at inference, under all allowable actions to estimate surprisal.
    \item Meaning of stochastic: The word stochastic implies some randomness within the setting. This randomness is derived from the possible set of all actions. In discriminative networks in Eq.~\ref{eq:Action_Dis}, $a_i, i \in [1,N]$ is the set of all possible actions with $N$ being the number of trained classes. This suggests that we allow a datapoint to be any available class, all of which are equally likely. Similarly, in generative networks in Eq.~\ref{eq:Action_Gen}, we add random perturbations at the output of the autoencoder. Hence, there is an inherent randomness within the actions that allow for the usage of the word \emph{stochastic}.
\end{enumerate}

\subsection{Abductive Reasoning}
\begin{figure*}[!htb]
\begin{center}
\minipage{1\textwidth}%
  \includegraphics[width=\linewidth]{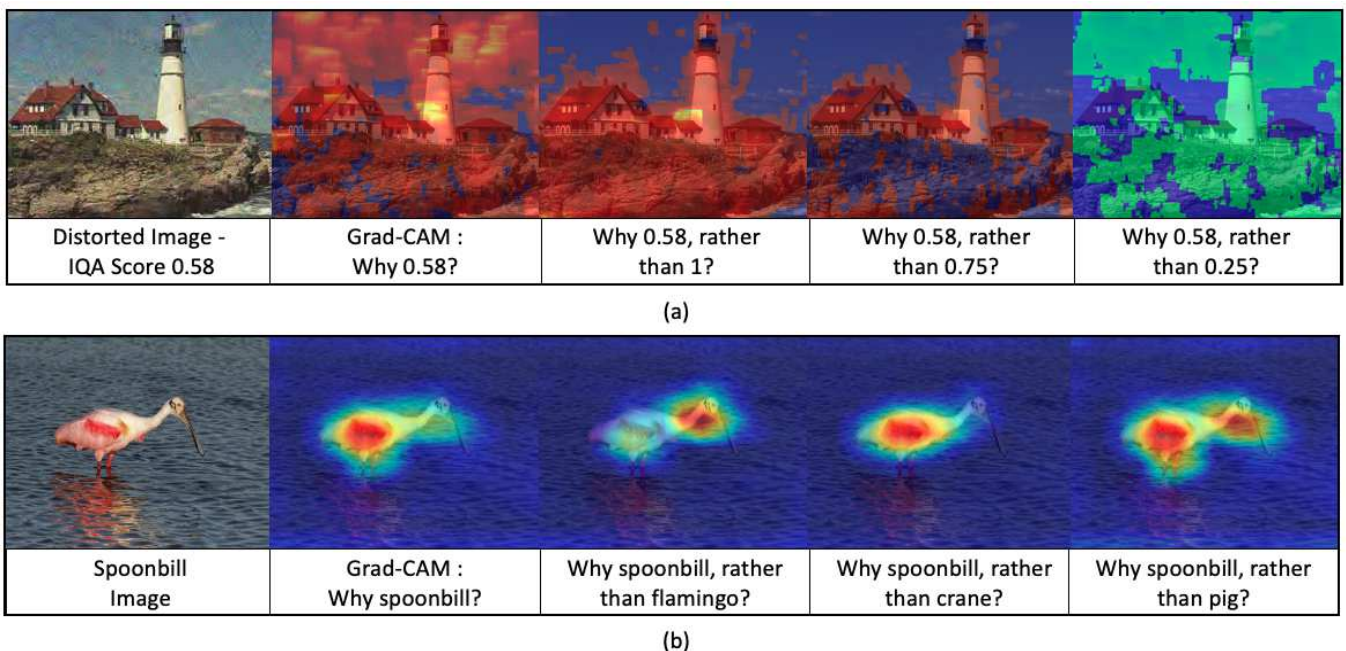}
\endminipage
\vspace{-3mm}
\caption{Stochastic surprisal answers contrastive questions. The highlighted regions in each image provides a visual explanation to the question beneath it. While Grad-CAM~\citep{selvaraju2017grad} shows all the perceived regions in the image, the stochastic surprisal provides fine-grained answers to contrastive questions. Best viewed in color.}\label{fig:IQA}
\vspace{-3mm}
\end{center}
\end{figure*}

The free energy principle postulates that the brain encodes a Bayesian recognition density that predicts sensory data based upon some hypotheses about their causes. This mode of inference is called inference to the best explanation. The underlying reasoning model is abductive reasoning. Abductive reasoning was introduced by the philosopher Charles Sanders Peirce~\citep{peirce1931collected}, who saw abduction as a reasoning process from effect to cause~\citep{paul1993approaches}. An abductive reasoning framework creates a hypothesis and tests its validity without considering the cause. A hypothesis can be considered as an answer to one of the three following questions: a causal \emph{`Why P?'} question, a counterfactual \emph{`What if?'} question, and a contrastive \emph{`Why P, rather than Q?'} question~\citep{alregib2022explanatory}. Here $P$ is the prediction and $Q$ is any contrast class. The action considered in this paper is the latter contrastive question of the form \emph{`Why P, rather than Q?'}. Stochastic surprisal measures the answer to this question. We explore this further in~\cite{alregib2022explanatory, prabhushankar2020contrastive}. We borrow the visualization procedure from~\cite{prabhushankar2020contrastive} to visually analyze stochastic surprise in the applications of IQA and recognition in Fig.~\ref{fig:IQA}. We do so to illustrate the broader impact of action at inference time. As in Section~\ref{subsec:IQA_Experiments}, we use stochastic surprisal as a plug-in approach. 

For IQA visualizations, we use a trained full-reference metric DIQaM-FR~\cite{bosse2017deep} as our perception model. In Fig.~\ref{fig:IQA}a, the pretrained network from~\cite{bosse2017deep} provides a quality score of $0.58$ to the distorted lighthouse image. Here $0.58$ acts as $P$ in the contrastive question. We use MSE loss function as $\mathcal{A}_d$ and a real number $Q \in [0,1]$ to calculate stochastic surprisal. Contrastive explanations of $Q$ values including $0.25, 0.75,\text{ and } 1$ along with Grad-CAM results are shown in Fig.~\ref{fig:IQA}a. Grad-CAM highlights the entire image indicating that the network estimates the quality based on the whole image. While this builds trust in the network, it does not help us understand the network decision. The stochastic surprisal, however, provides fine-grained explanations. Consider the contrastive questions asking why the quality is neither  $1\text{ nor } 0.75$. The network estimates this to be primarily due to distortions concentrating in the foreground portion of the image. This explanation is inline with previous works in IQA that posit that distortions in the more salient foreground or edge features cause a larger drop in perceptual quality than that in color or background~\citep{prabhushankar2017ms}\citep{chandler2013seven}. When the contrastive question asks why the prediction is not $0.25$, the network highlights the sky indicating its good quality for a higher score of $0.58$. 

Fig.~\ref{fig:IQA}b shows the contrastive questions answered by the stochastic surprisal for the application of recognition. Given an image of a spoonbill from ImageNet dataset~\citep{deng2009imagenet}, a VGG-16 network highlights the body, feathers, legs and beak of the bird in the Grad-CAM~\citep{selvaraju2017grad} explanation. Consider a more fine grained contrastive question regarding the difference between a spoonbill and flamingo. The stochastic surprisal highlights regions in the neck of the spoonbill indicating that the contrast between the input spoonbill image and the network's notion of a flamingo lies in the spoonbill's lack of S-shaped neck. Similarly, the contrast between a spoonbill and a crane is in the color of the spoonbill's feathers. The contrast between a pig and a spoonbill is in the shape of neck and legs in the spoonbill which is emphasized. All these visualizations serve to illustrate the stochastic nature of the proposed method. It is stochastic in the sense that it individually depends on the network, the data, as well as the action. In this case, the action of not predicting a flamingo has a different explanation compared to the action of not predicting a pig.

\subsection{Expectancy-Mismatch}
The expectancy-mismatch hypothesis in cognitive science is a way to quantify and analyze human attention. According to this hypothesis, human attention mechanism suppresses expected messages and focuses on the unexpected ones~\citep{SUMMERFIELD2009403,krebs2012stimulus,horstmann2016perceptual,doi:10.1111/1467-9280.00488,BECKER2011290,sun2020implicit}.~\cite{BECKER2011290} shows that a message which is unexpected, captures human attention. Then, the human visual system establishes whether the input matches the observers' expectation. If they are conflicting, error neurons in the human brain encode the prediction error and pass the error message back to the representational neurons. The proposed method uses gradients with respect to the network parameters to measure an action. In both the generative and discriminative networks, this action takes the form of a change in the output thereby creating a mismatch with the network's expected result. Hence, the proposed method can act as a framework for exploring expectancy-mismatch in future works.

\subsection{Related Learning Paradigms}
The proposed stochastic surprisal decomposes the decision making and training process of a neural network into perception and action phases. A number of other machine learning paradigms including continual and lifelong learning~\citep{parisi2019continual}, online learning~\citep{hoi2021online}, and introspective learning~\citep{prabhushankar2022introspective} also have multiple stages. Online learning assumes an exploration and exploitation stage in a neural network's training process. Hence, the differentiation in the training stages is based on time rather than the proposed action. Continual and lifelong learning is a research paradigm that tackles the topic of catastrophic forgetting when a neural network is trained to perform multiple tasks. Introspective learning conjectures reasons in the form of counterfactual or contrastive questions in its two stages to make predictions. Hence, while there are multiple machine learning paradigms that conjecture decomposition of neural network's training and decision processes, the proposed framework that is based on the FEP is unique in its decomposition. The field of active learning~\citep{logan2022decal, benkert2022forgetful} involves actions within the training and decision making processes. However, active learning requires actions from the users while the considered actions in the proposed methodology are with respect to the neural network.

\subsection{Conclusion}
\label{subsec:Future}
In this paper, we examine supervised learning from the perspective of Free Energy Principle. The learning process of both generative and discriminative models can be decomposed into divergence and surprisal measures. Surprisal is introduced in generative models via regularization and constraints that allow a generative aspect to their functionality. While this complicates the action itself, the set of possible actions is still limited. Discriminative networks follow the traditional route of free energy minimization by defining surprisal in terms of recognition entropy and minimizing it. This allows the action itself to be a simple fidelity-based reconstruction error. However, in discriminative networks, there are $N$ set of possible actions, $N$ being the number of classes in the recognition density. We account for both these peculiarities in defining our action space. We use a fidelity-based MSE loss for both generative and discriminative networks. In addition, generative networks are reinforced with KL-divergence based elastic net regularization, and in discriminative networks we backpropagate $N$ possible actions. We measure this scalar action quantity in terms of a vector quantity called stochastic surprisal that is a function of the network parameters and an individual data point rather than a distribution. We use stochastic surprisal to assess distortions in image quality assessment and disregard distortions in robust recognition.  We then discuss the implications of stochastic surprisal in other areas of cognitive science including abductive reasoning and expectancy-mismatch. A computational bottleneck within the framework is the consideration of all $N$ possible actions to estimate the surprisal feature $r_x$. $r_x$ scales linearly with $N$ thereby becoming prohibitive on datasets with a large number of classes. Selecting only a subset of the most likely actions is one plausible solution to the challenge of scalability.